\documentclass{article}
\usepackage{times}

\usepackage{graphicx}
\usepackage{caption}
\usepackage{subcaption}
\usepackage{tikz}
\usepackage{pgfplots}

\usepackage{multirow}
\usepackage{multicol}

\usepackage{amsmath}
\usepackage{amssymb}
\usepackage{amsfonts}
\usepackage{bbm}
\usepackage{bbold}

\usepackage{algorithm}
\usepackage{algpseudocode}
\usepackage{setspace}
\usepackage{mathtools}

\usepackage{hyperref}
\usepackage{cleveref}
\usepackage{natbib}

\usepackage{enumitem}

\usepackage{verbatim}

\newcommand{\model}{f}
\newcommand{\myxspace}{\mathcal{X}}
\newcommand{\yspace}{\mathcal{Y}}
\newcommand{\Noisyy}{\widetilde{Y}}
\newcommand{\noisyy}{\widetilde{y}}
\newcommand{\prob}{\mathbb{P}}
\newcommand{\expectation}{\mathbb{E}}
\newcommand{\entropy}[1]{\mathcal{H}[#1]}

\newcommand{\logt}{\log_2}
\newcommand{\argmax}[1]{\underset{#1}{\text{argmax}}\;}

\newcommand{\X}{X}
\newcommand{\Y}{Y}
\newcommand{\fullint}{\int_{-\infty}^{\infty}}

\newcommand{\gaussian}{\mathcal{N}}
\newcommand{\acqfun}{\phi_{\model}}
\newcommand{\pooldata}[1]{X_{1:#1}^{(\text{pool})}}
\newcommand{\datatrain}[1]{\mathcal{D}^{(\text{train})}_{#1}}
\newcommand{\noisydatatrain}[1]{\widetilde{\mathcal{D}}^{(\text{train})}_{#1}}
\newcommand{\xtrain}[1]{X_{1:#1}^{(\text{train})}}
\newcommand{\ytrain}[1]{Y_{1:#1}^{(\text{train})}}
\newcommand{\atrain}[1]{\alpha_{1:#1}^{(\text{train})}}

\newcommand{\noisyytrain}[1]{\widetilde{Y}_{1:#1}^{(\text{train})}}

\newcommand{\xselect}{X^{(\text{a})}}

\newcommand{\aselect}{\alpha^{(\text{a})}}

\newcommand{\alphaset}{\mathcal{A}}

\newcommand{\mucavity}{\mu_{-i}}
\newcommand{\sigmacavity}{\sigma_{-i}}

\usepackage{tikz}
\usetikzlibrary{shapes.geometric, arrows, positioning, arrows.meta, calc, patterns, shapes.misc, matrix, backgrounds, external}
\usepackage{pgfplots}
\usepgfplotslibrary{groupplots}

\usepackage{caption}
\usepackage{subcaption}

\newif\ifarxiv
\arxivtrue

\newcommand{\authorliu}[2]{
{\normalsize{\bf #1}}\\
\normalsize{#2@liu.se} \\
\normalsize{Link\"oping University}
}

\newcommand{\authorchalmers}[2]{
{\normalsize{\bf #1}}\\
\normalsize{#2@chalmers.se} \\
\normalsize{Chalmers University of Technology}
}

\title{Active Learning with Weak Supervision\\ for Gaussian Processes}

\author{
\authorliu{Amanda Olmin}{amanda.olmin}
\and 
\authorchalmers{Jakob Lindqvist}{jakob.lindqvist}
\and
\authorchalmers{Lennart Svensson}{lennart.svensson}
\and
\authorliu{Fredrik Lindsten}{fredrik.lindsten}
}

\date{}

\begin{document}

\maketitle

\begin{abstract}
    Annotating data for supervised learning can be costly. When the annotation budget is limited, active learning can be used to select and annotate those observations that are likely to give the most gain in model performance. We propose an active learning algorithm that, in addition to selecting which observation to annotate, selects the precision of the annotation that is acquired. Assuming that annotations with low precision are cheaper to obtain, this allows the model to explore a larger part of the input space, with the same annotation budget. We build our acquisition function on the previously proposed BALD objective for Gaussian Processes, and empirically demonstrate the gains of being able to adjust the annotation precision in the active learning loop.
\end{abstract}

\section{Introduction}
Supervised learning requires annotated data and sometimes a vast amount of it. In situations where input data is abundant or the input distribution known, but annotations are costly, we can use the annotation budget wisely and optimise model performance by selecting and annotating those instances that are most useful for the model. This is often referred to as \textit{active learning}. So called pool-based active learning typically adopts a greedy strategy where instances are iteratively selected and added to the training set using a fixed acquisition strategy  \citep{Settles2009}. 
However, while the focus of most active learning algorithms lies on instance selection, there might be other factors that can be controlled in order to use the annotation budget optimally. 

In this paper, we identify the precision of annotations as a factor that can help improve model performance in the active learning algorithm. Under circumstances where it is possible to obtain cheaper, but noisier, annotations, gains can be made in model performance by collecting several noisy annotations, in place of a few precise. Consider, as an example, applications where annotations are acquired through expensive calculations or simulations, 
and where the processing time can be used to tune the numerical precision. If less precise annotations are acquired, corresponding to faster processing, it could allow the model to explore a larger part of the input space with the given annotation budget, compared to if only annotations of high precision are used. 

In other situations, being able to control the precision of annotations can help in using resources more wisely. For instance, if annotations are obtained through empirical experiments and the precision is determined by the number of repeated experiments, time and material could be saved if the number of experiments is determined beforehand. This, as it would allow experiments to be performed group-wise instead of sequentially. 

With the given motivation in mind, we propose an extended, iterative active learning algorithm that, in addition to instance selection, optimise for the precision of annotations. We refer to this method as active learning with \textit{weak} supervision. 
The proposed acquisition function is inspired by Bayesian Active Learning by Disagreement (BALD, \cite{Houlsby2011}) and is based on the mutual information between the weak annotation and the model. In each iteration of the active learning algorithm, we optimise mutual information per annotation cost. Hence, the proposed acquisition function can be interpreted as conveying which annotations that will give the most information about the model per cost unit. 

\section{Related work}
The proposed setting, where annotation precision and cost are included in the active learning loop, is relatively under-explored in active learning. However, a similar approach to the one proposed in this paper is introduced in \citep{Li2022}, but with some important differences. Firstly, we propose an acquisition function based on the mutual information between the weak annotation and the model, instead of the target variable as in \citep{Li2022}, and that does not require any design choices for the latent target variable. Secondly, our method is easily adapted to any model that can account for weak annotations, and is not restricted to one type of annotation error, or noise, or one type of task. In contrast, the method in \citet{Li2022} is adapted to the setting of a high-dimensional model output, where precision is determined by the mesh size of the target variable, and focuses on regression. 

A special case of the active learning algorithm with weak supervision is that of selecting the most appropriate annotator, considering both annotator accuracy and cost. 
Most closely related to the algorithm proposed in this paper, in this context, are iterative active learning algorithms that simultaneously select instances and annotators; mainly those that specifically consider annotator costs,  e.g. \citep{Huang2017,Gao2020,Chakraborty2020}, but also those that do not, e.g. \citep{Yan2011,Herde2020}. 

Closely related to the proposed active learning algorithm, is the use of multi-fidelity function evaluations in Bayesian optimisation, e.g. \citep{Picheny2013,Song2019,Li2020,Takeno2020}, and design optimisation, e.g. \citet{Pellegrini2022}. 
In contrast to active learning, the goal of these related methods is to find the optimum of a function and not to optimise model performance. Related to the algorithm presented in this paper is also that of finding high-accuracy metamodels in a multi-fidelity setting \citep{Wu2020,Tian2020}. However, to the best of our knowledge, this branch of work considers only situations including two fidelity levels, while we allow any number of fidelity, or precision, levels.

\section{Active learning with weak supervision}
Suppose that we want to learn a probabilistic predictive model, $\model: \myxspace \rightarrow \yspace$, predicting the distribution of a target variable $Y \in \yspace$ given the input variable $X \in \myxspace$. In the pool-based active learning setting we either have a set of $N$ observations, or know the $N$ possible values, of the input variable, denoted by $\pooldata{N}$. For the purpose of using supervised learning, we need to collect corresponding target values. However, our annotation budget, $B$, is limited.

To select which instances to annotate and use for training in order to optimise model performance, we consider iterative active learning. A traditional active learning algorithm of this kind focuses on instance selection. However, our proposed algorithm is based on the argument that we can improve model performance further by allowing to control the precision of annotations, denoted by $\alpha$. We assume that $\alpha$ belongs to a set of precision levels, $\alphaset$, and controls the distribution of a weak, or noisy, annotation variable $\Noisyy \in \yspace$ that we observe in place of $\Y$.  For instance, $\alpha$ could control the variance of $\Noisyy$. A higher precision corresponds to a more accurate annotation, typically meaning that the properties of $\Noisyy$ are closer to those of $\Y$. 
Our proposed algorithm, active learning with weak supervision, repeats the following steps until the budget is exhausted:
\begin{enumerate}
    \item Fit the model, $\model$, to the current set of annotated data.
    \item Based on the current model $\model$, select the next instance to annotate \textbf{and the precision of the annotation}.
    \item Annotate the selected instance \textbf{with the selected precision} and add the new data pair to the training data set.
\end{enumerate}
The aspects that differ from the traditional active learning algorithm have been marked in bold. Typically, the active learning procedure starts with an initial training set of $n$ annotated observations, $\noisydatatrain{n} = (\xtrain{n}, \noisyytrain{n}, \atrain{n})$, which is then gradually expanded. 

The training set that is collected using active learning with weak supervision, contains weak annotations. Since the precision of an annotation is known, we include it in the learning process, to account for e.g. additional noise in the data. For this purpose, we specify a generative model of $\Noisyy$, illustrated by the graphical model in \cref{fig:pgm_noise}(a), where $\Noisyy$ is dependent on both $\X$ and $\model$. This generative model will also allow for evaluating the proposed acquisition function introduced below. 


%
%

\subsection{Acquisition functions for precision selection}
For the selection step of the active learning algorithm with weak supervision, we define an acquisition function, $\acqfun$, that conveys our intention of selecting the instance, $\xselect \in \pooldata{N}$, and annotation precision, $\aselect$, that will give the most gain in model performance given the current model, $\model$. The selection is performed by solving the following optimisation problem
\begin{align}
    \xselect, \aselect = \argmax{X \in \pooldata{N}, \alpha \in \alphaset} \acqfun(X, \alpha),
\end{align}
%

We build our acquisition strategy on the Bayesian Active Learning by Disagreement (BALD, \citet{Houlsby2011}) acquisition function, which is defined as the mutual information (MI) between the target variable, $\Y$, and the model, $\model$, conditioned on the input as well as the training data $\datatrain{n}=(\xtrain{n}, \ytrain{n})$ 
\begin{align}
    \text{MI}(\Y;\model \mid \X, \datatrain{n}) = &\entropy{\Y \mid \X, \datatrain{n}}  - \expectation_{\model \mid \X, \datatrain{n}}\big[\entropy{\Y \mid \X, \model}\big].
    \label{eq:mi_bald}
\end{align}
Here, $\entropy{\cdot}$ refers to the entropy of a random variable \citep{Thomas2006}. 

%
%
%
To optimise mutual information and at the same time account for annotation costs, in the greedy fashion of iterative active learning, we optimise information per cost. To this end, we introduce the cost function $C(\alpha)$, describing the cost of acquiring an annotation with precision $\alpha$. We put no limitations in regards to which cost function is used, as long as it can be properly evaluated. In practice, it should be adapted to the application, such that it corresponds to the actual cost, or time, needed to annotate a data point with a certain precision. The proposed acquisition function, adapted to the setting with weak annotations, is
\begin{align}
    \acqfun(X, \alpha) = \text{MI}(\Noisyy;\model \mid \X, \alpha, \noisydatatrain{n}) / C(\alpha)
    \label{eq:prop_obj}
\end{align}
Hence, we want to select the instance, and corresponding annotation precision, for which the mutual information between the weak annotation, $\Noisyy$, and the model, $\model$, is high, while at the same time keeping the annotation cost low. We will, for short, refer to this acquisition function by $\text{MI}(\Noisyy;\model)$. 

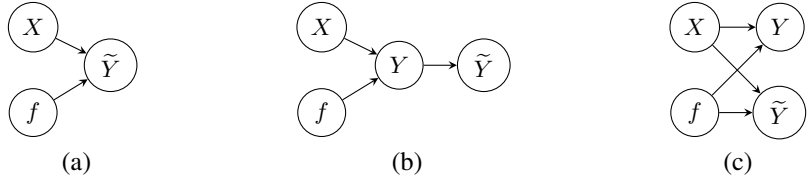
\begin{figure}[t]
    \centering
    \newcommand{\equal}{=}
\newcommand{\dist}{0.8}

\ifarxiv
\newcommand{\sepdist}{3.5}
\newcommand{\sepdistt}{4.5}
\else
\newcommand{\sepdist}{2.5}
\newcommand{\sepdistt}{3.5}
\fi
\newcommand{\labdist}{1.7}
\newcommand{\minw}{0.5}
\fontsize{9}{8}\selectfont

\begin{tikzpicture}[scale=0.9]
    
    \node[circle, minimum width=\minw cm, draw=black, anchor=center] (x1) at (0, 0) {$X$};
    \node[circle, minimum width=\minw cm, draw=black, below=\dist cm of x1, anchor=center] (f1) {$\model$};
    \node[circle, minimum width=\minw cm, draw=black, above right=0.5*\dist cm and \dist cm of f1, anchor=center] (yt1) {$\Noisyy$};
    
    \draw[black, -stealth] (x1) -- (yt1);
    \draw[black, -stealth] (f1) -- (yt1);
    
    \node[right=0.3*\dist cm of x1, anchor=center] (tanch1) {};
    \node[below=\labdist cm of tanch1, anchor=center, align=left] (t1) {\normalsize{(a)}};
        
     \node[circle, minimum width=\minw cm, draw=black, right = \sepdist cm of x1, anchor=center] (x3) {$X$};
    \node[circle, minimum width=\minw cm, draw=black, below = \dist cm of x3, anchor=center] (f3) {$\model$};
    \node[circle, minimum width=\minw cm, draw=black, above right=0.5*\dist cm and \dist cm of f3, anchor=center] (y3) {$Y$};
    \node[circle, minimum width=\minw cm, draw=black, right= \dist cm of y3, anchor=center] (yt3) {$\Noisyy$};
    
    \draw[black, -stealth] (x3) -- (y3);
    \draw[black, -stealth] (f3) -- (y3);
    \draw[black, -stealth] (y3) -- (yt3);

    \node[right=\dist cm of x3, anchor=center] (tanch3) {};
    \node[below=\labdist cm of tanch3, anchor=center, align=left] (t3) {\normalsize{(b)}};

    \node[circle, minimum width=\minw cm, draw=black, right = \sepdist + \sepdistt cm of x3, anchor=center] (x2) {$X$};
    \node[circle, minimum width=\minw cm, draw=black, below = \dist cm of x2, anchor=center] (f2) {$\model$};
    \node[circle, minimum width=\minw cm, draw=black, right= \dist cm of x2, anchor=center] (y2) {$Y$};
    \node[circle, minimum width=\minw cm, draw=black, right= \dist cm of f2, anchor=center] (yt2) {$\Noisyy$};
    
    \draw[black, -stealth] (x2) -- (y2);
     \draw[black, -stealth] (f2) -- (y2);
    \draw[black, -stealth] (x2) -- (yt2);
    \draw[black, -stealth] (f2) -- (yt2);
    
    \node[right=0.3*\dist cm of x2, anchor=center] (tanch2) {};

    \node[below=\labdist cm of tanch2, anchor=center, align=left] (t2) {\normalsize{(c)}};

\end{tikzpicture}
    \caption{Generative models for the weak target variable, $\Noisyy$. (a) Generative model without $\Y$. (b) $\Noisyy$ is conditionally independent of $\model$ and $\X$ given $\Y$. (c) $\Noisyy$ and $\Y$ are independent given $\model$ and $\X$.}
    \label{fig:pgm_noise}
\end{figure}

As an alternative to \cref{eq:prop_obj}, we also consider the mutual information between the weak annotation and the target variable, replacing $\model$ with $\Y$ in \cref{eq:prop_obj}, as in \citep{Li2022}. For abbreviation, we will use $\text{MI}(\Noisyy; \Y)$ to refer to this acquisition function. Although not necessary for learning the model, the $\text{MI}(\Noisyy;\Y)$ acquisition function requires that we specify a generative model of the latent variable, $\Y$. The most natural design choice will depend on the application. We could imagine, for instance, that $\Y$ represents a true target value and that $\Noisyy$ is a noisy version of this true target, illustrated by the graphical model in \cref{fig:pgm_noise}(b). An alternative is to regard both $\Y$ and $\Noisyy$ as noisy, independent measurements, or observations, of the model output $\model(X)$, as illustrated in \cref{fig:pgm_noise}(c).

%
%

\section{Gaussian Processes with weak annotations}
A Gaussian Process (GP) is a probabilistic, non-parametric model, postulating a Gaussian distribution over functions, $\model: \myxspace \rightarrow \mathbb{R}$, see e.g. \citet{Rasmussen2006}. Although GPs can be extended to the multivariate setting, we will introduce our model for the case in which $\yspace \subseteq \mathbb{R}$. Then, we define the Gaussian Process prior as 
\begin{align}
    &\model \sim \text{GP}(m(\cdot), K(\cdot, \cdot))
    \label{eq:gp_prior}
\end{align}
where $m(\cdot)$ and $K(\cdot, \cdot)$ are mean value and kernel functions, respectively. We will assume that $m(\cdot)$ is $0$. An example of a kernel is the commonly used RBF kernel (see e.g. \citet{Rasmussen2006}): $K(X_i, X_j) = a^2 \exp\left(-2l^{-2}\|X_i-X_j\|_2^2\right)$, where $\|\cdot\|_2$ is the Euclidean norm and $a$, $l$ are hyperparameters.

We further assume, in line with the standard Gaussian Process regressor, that the conditional distribution of $\Noisyy$ is Gaussian 
\begin{align}
    \Noisyy \mid \model, \X, \alpha \sim \gaussian \big(\model(\X), \sigma^2(X) + \gamma/\alpha \big).
    \label{eq:regr_noise_model}
\end{align}
The precision, $\alpha$, controls the variance of the weak annotation variable. We will assume that $\alpha \in \alphaset \subseteq [1.0, \infty)$, such that the smallest attainable variance, possibly depending on the input, is $\sigma^2(X)$. The lowest precision, in contrast, gives the maximum variance, $\sigma^2(X) + \gamma$. The parameter $\gamma > 0$ is a constant. 
%
%
%
%
%
%
%

Based on the introduced probabilistic model, we can derive the predictive distribution given a new observation $X$ and precision $\alpha$ to find
%
%
%
\begin{align}
    &\Noisyy \mid \X, \alpha, \noisydatatrain{n} \sim \gaussian \big(\widetilde{\mu}_*, \widetilde{\sigma}_*^2 + \sigma^2(X) + \gamma/\alpha \big), 
    \label{eq:gp_pred_ws}
\end{align}
where the parameters $\mu_*$ and $\sigma_*^2$ have closed form expressions and will depend on the training data $\noisydatatrain{n}$ used to learn the model, see e.g. \citep{Rasmussen2006}.
%
%

Following the predictive distribution, and the expression for the differential entropy of a Gaussian random variable, the denominator of \cref{eq:prop_obj} evaluates to
\begin{align*}
    \text{MI}(\Noisyy;\model \mid \X, \alpha, \noisydatatrain{n}) = 0.5\big(\log \big(\widetilde{\sigma}_*^2 + \sigma^2(X) + \gamma/\alpha \big) - \log \big(\sigma^2(X) + \gamma/\alpha \big)\big).
\end{align*}

For evaluating $\text{MI}(\Noisyy; \Y)$, we will assume that $\Y$ follows the distribution in \cref{eq:regr_noise_model} with $\alpha \rightarrow \infty$. As discussed, this acquisition function will also depend on how we model the latent variable $\Y$ in terms of its relation to the other variables.

\subsection{Gaussian Process classifier}
For classification, we consider the binary setting with $\yspace = \{-1, 1\}$. The Gaussian Process classifier is obtained by replacing \cref{eq:regr_noise_model} of the GP regressor with
\begin{align}
    &\Noisyy \mid \model, X, \alpha \sim \text{Bernoulli} \big((2\omega_\alpha - 1)\Phi(\model) + 1 - \omega_\alpha) \big), \quad \omega_\alpha = \kappa + \gamma \alpha
    \label{eq:gpc_prior}
\end{align}
where $\Phi(\cdot)$ is the standard Gaussian cdf. We assume symmetric, input-independent label noise, following the graphical model in \cref{fig:pgm_noise}(b). The label flip probability, $1-\omega_\alpha$, depends on the precision, $\alpha$, as well as the constants $\kappa$ and $\gamma$. 
We will assume that $\alpha \in \alphaset = [0.0, 1.0]$, such that $1-\omega_\alpha \in [1 - (\kappa + \gamma), 1 - \kappa]$. 

The posterior of the GP classifier is non-Gaussian and intractable, but is typically approximated by a Gaussian distribution. In the experiments, we approximate the posterior over $\model$ using expectation propagation (EP), following \citet{Rasmussen2006}, after which we can estimate the parameters of the approximate predictive distribution $\model \mid \X,  \noisydatatrain{n} \sim \mathcal{N}\big(\mu_*, \sigma_*^2)$. 
%

%
%
%

The conditional mutual information between $\Noisyy$ and $\model$ for binary classification is given by
\small{
\begin{align*}
    &\text{MI}(\Noisyy;\model \mid \X, \alpha, \noisydatatrain{n}) \approx h\bigg( \Phi \Big( \frac{\mu_*}{\sqrt{\sigma_*^2 + 1}} \Big) \bigg)   - \frac{\big(1-h(\omega_\alpha)\big)}{\sqrt{1 + 2C\sigma_*^2}}\exp\bigg(\frac{-C \mu_*^2}{1 + 2C\sigma_*^2} \bigg) + h(\omega_\alpha), 
\end{align*}}
\normalsize
\hspace{-0.6em}where $\quad C = (2\omega_\alpha - 1)^2 (\pi \log(2) (1 - h(\omega_\alpha)))^{-1}$ and the function $h(\cdot)$ is the Shannon entropy. Similar to \citep{Houlsby2011}, the second term in the expression is approximated using a Taylor expansion of order three, but of the function $g(x) = \log \big( h((2\omega_\alpha - 1)\Phi(x) + 1 - \omega_\alpha) - h(\omega_\alpha) \big)$.

To evaluate the $\text{MI}(\Noisyy;\Y)$ acquisition function, we will assume that $\Y$ follows the distribution in \cref{eq:gpc_prior} with $\alpha=1.0$.

%
%

\section{Experiments}
The proposed acquisition function, $\text{MI}(\Noisyy; \model)$, is compared to the $\text{MI}(\Noisyy; \Y)$ and BALD acquisition functions, as well as to randomly sampling data points from $\pooldata{N}$ with maximum annotation precision. We use GP models with RBF kernels, where hyperparameters are set as $a=l=1.0$, if nothing else is mentioned. We run all experiments 15 times and report the first, second (median) and third quartiles of the performance metric as a function of the total annotation cost. In cases where the number of data points differ between experiments, we interpolate the results and visualise the corresponding curves without marks.\footnote{Code provided at \url{https://github.com/AOlmin/active_learning_weak_sup}}

For $\text{MI}(\Noisyy; \model)$ and $\text{MI}(\Noisyy; \Y)$, when $\alphaset$ is continuous, we perform optimisation with respect to $\alpha$ by making a discretisation over $\alphaset$, as the optimisation problem can typically not be solved analytically. %

Although we resort to this simple solution, it is also possible to directly optimise the acquisition function over a continuous interval, using a numerical optimisation method of choice. 

\paragraph{Sine curve.} In the first set of experiments, we generate data as
\begin{align}
    &\Noisyy \mid X, \alpha \sim \mathcal{N}\Big(0.2 X \cdot \sin(\omega X), 0.01 \big(1 + \big(X/5.0\big)^2\big) + 0.09/\alpha \Big),
    \label{eq:sine_toy_data}
\end{align}
where $\X$ is sampled uniformly from $\myxspace = [0.0, 5.0)$ and with $\omega=3.0$, if nothing else is mentioned. The variance of $\Noisyy$ given $\model(X)$ is a maximum of ten times as large as the conditional variance of $\Y$, which has a precision of $\alpha \rightarrow \infty$. For evaluating $\text{MI}(\Noisyy;\Y)$, we initially follow the graphical model in \cref{fig:pgm_noise}(c). For each experiment, we sample a data set of 8,000 data points, whereof 75\% is allocated to the data pool and the remaining 25\% to a test set. For the initial training set, $n=10$ data points are randomly sampled from the data pool and annotated with maximum precision. 

\begin{figure}[t]
    \includegraphics[width=\textwidth]{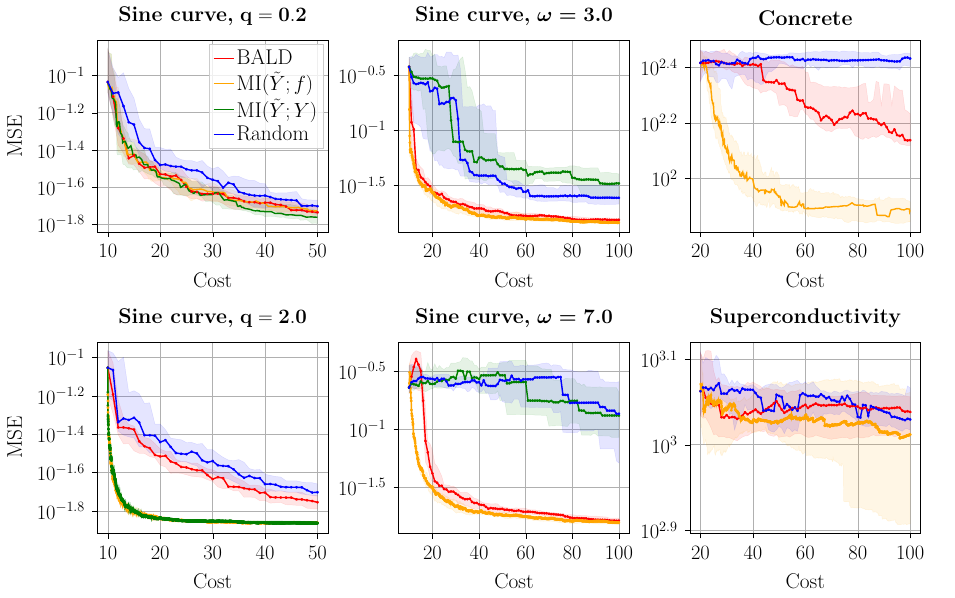}
    \caption{Median, first and third quartiles of the test MSE obtained from each set of experiments. \textit{Left}: Sine curve experiments using cost functions with varying parameter $q$. \textit{Middle}: Sine curve experiments with an under-explored input space and where $\Noisyy$ is independent of $\model$ and $\X$ given $\Y$. \textit{Right}: UCI data experiments. 
    }
    \label{fig:regr_exp}
\end{figure}

We perform experiments with cost functions of the form $C(\alpha) = \big(1 + c\alpha^{-1}\big)^{-q}$, where we set $c = \gamma/\kappa$, such that the cost of annotating with precision $\alpha$ is inversely proportional to the variance of $\Noisyy$. The parameter $q$ will additionally control the relative cost of annotating with low precision. The test Mean Squared Error (MSE), using a budget of $B=50$, for two different values of $q$, namely $q=0.2$ and $q=2.0$, is reported in the left column of \cref{fig:regr_exp}. The most gain with adjusting the annotation precision is obtained with $q=2.0$. In this case, it is much cheaper to acquire noisy annotations than annotations of high precision, and the proposed algorithm consistently selects the lowest precision. In contrast, when $q=0.2$, a majority of the annotations are acquired with maximum precision and $\text{MI}(\Noisyy; \model)$ behaves similarly to standard BALD. 

We next run active learning experiments in a setting where part of the input space is under-explored. Observations in the data pool are sampled with probability 0.9 from the first half of the input space, $[0.0, 2.5)$, and from the second half, $[2.5, 5.0)$, otherwise. We still evaluate model performance on the full input space and therefore sample the observations in the test set uniformly. We argue that circumstances like these are not uncommon in practice. For example, it could be important that the model performs well also on rare observations or we could have a bias in the data collection process. In the experiments that follow, we also assume that $\Noisyy$ is generated from $\Y$ and obeys the distribution $\mathcal{N}(Y, 0.09/\alpha)$. 

We perform two set of experiments with the aforementioned setting, using $\omega=3.0$ and $7.0$, respectively, in \cref{eq:sine_toy_data}. The length scale of the RBF kernel is adjusted as $l=3.0/\omega$. We use a budget of $B=100$ and cost function defined as above with $q=1.0$. Results are shown in the middle column of \cref{fig:regr_exp}. The $\text{MI}(\Noisyy;\model)$ and BALD acquisition functions give significantly better model performance than random sampling. Moreover, as the function frequency, $\omega$, is increased, the advantage of adjusting the precision of annotations gets more apparent, likely because it is extra beneficial to be able to explore the input space when the frequency is high. The reason for the poor performance of $\text{MI}(\Noisyy;\Y)$ is that the mutual information between a continuous random variable and itself is infinite. The acquisition function is therefore independent of $\X$ for $\alpha \rightarrow \infty$. 

\paragraph{UCI data sets.} We test the proposed active learning algorithm on the concrete compressive strength \citep{Yeh1998} and superconductivity \citep{Hamidieh2018} data sets from the UCI Machine Learning Repository \citep{Dua2019}. In both cases, we add artificial, input-independent noise by sampling $\Noisyy$ from the distribution $\mathcal{N}(Y, 1/\alpha)$. We assume that $\Y$ is known, with high precision, from the data and set $\sigma^2(X)=1\cdot 10^{-3}$.  For each experiment, we make a 80\%-20\% pool-test split, and sample an initial training set of size $n=20$ from the data pool. Hyperparameters of the models' RBF kernels are fitted using maximum likelihood optimisation. 

We run experiments using a budget of $B=100$ and the same cost function as above with $q=1.0$ and $c=9.0$, such that annotating with the lowest precision costs one tenth of annotating with $\alpha \rightarrow \infty$. Results are shown in the right column of \cref{fig:regr_exp}. $\text{MI}(\Noisyy;\Y)$ has been excluded because of the poor performance in the previous experiments. Active learning, and $\text{MI}(\Noisyy;\model)$ in particular, improves model performance in both cases, but especially on the concrete data set.

\begin{figure}[t]
    \includegraphics[width=\textwidth]{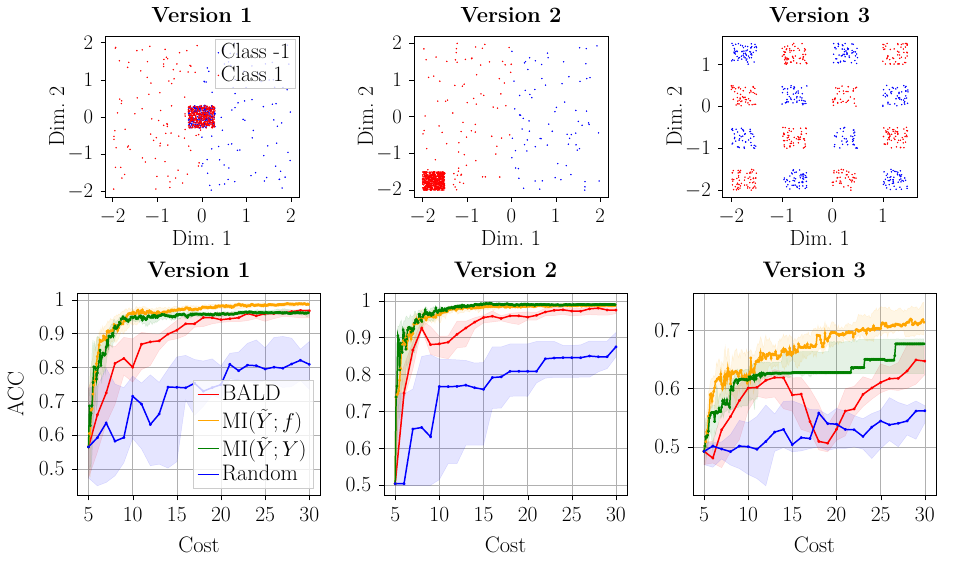}
     \caption{\textit{Top}: Examples of each of the three artificial classification data sets. Negative labels are given in blue and positive in red. \textit{Bottom}: Median, first and third quartiles of the test accuracy obtained from each set of experiments. 
    }
    \label{fig:toy_class_exp}
\end{figure}

\paragraph{Classification.} We next perform experiments with binary classification, generating data sets similar to the three artificial ones used in \citep{Houlsby2011}. The input variable in all data sets lies within a block ranging from -2.0 to 2.0 in two dimensions. For the first two data sets, there is a true classification boundary at zero in the first dimension, but one (Version 1) has a block of noisy labels at the decision boundary, while the other (Version 2) has a block of uninformative samples in the positive class. The third data set (Version 3) has classes organised in a checker-board pattern. Examples are shown in \cref{fig:toy_class_exp}. Data sets are generated with $8,000$ data points with a 75\%-25\% pool-test split, and an initial training set size of $n=5$. We set $\kappa = 0.8$ and $\gamma = 0.2$ in \cref{eq:gpc_prior}. 

Experiments are performed using a budget of $B=30$ and a linear cost function of the form $C(\alpha) = b + c\alpha$. The parameters of the cost function are set such that annotations with the highest precision has a cost of one, while the cost of annotating with the lowest precision is a tenth of that, with $b=0.1$ and $c=0.9$. Results are shown in \cref{fig:toy_class_exp}. Model performance improves in all cases when allowing for the use of weak annotations. Moreover, an advantage of $\text{MI}(\Noisyy; \model)$ over $\text{MI}(\Noisyy; \Y)$ is observed, particularly for the checker-board data set. Examples of factors affecting the success of using weak annotations are the maximum label flip probability and the cost function. 

\section{Conclusion}
We introduced an extension of active learning for Gaussian Processes, that includes the precision of annotations in the selection step. The proposed acquisition function is based on the mutual information between the weak annotation and the model, and does not require a generative model of the latent target variable. We demonstrated empirically how active learning with weak supervision can give a performance advantage in situations where it is cheaper to obtain several weak, in place of one or a few precise, annotations. When using weak annotations, the model can explore the input space to a larger degree than what is allowed by the budget if only annotations of high precision are acquired. For future work, we could investigate alternatives to optimising the acquistion function with respect to the annotation precision by discretising the set of precision levels, and extend the algorithm to performing several queries per iteration.

\section*{Acknowledgements}
This research is financially supported by the Swedish Research Council (project 2020-04122),
the Wallenberg AI, Autonomous Systems and Software Program (WASP) funded by the Knut and Alice Wallenberg Foundation,
and
the Excellence Center at Linköping--Lund in Information Technology (ELLIIT). This preprint has not undergone peer review 
or any post-submission improvements or corrections. The Version of Record of this contribution is published in Communications in Computer and Information Science, vol 1792, and is available online at https://doi.org/10.1007/978-981-99-1642-9\_17.

\medskip
\bibliographystyle{apalike}
\bibliography{main}
\newpage

\def\thesection{\Alph{section}}
\setcounter{section}{0}

\section*{Supplementary material for \textit{Active Learning with Weak Supervision for Gaussian Processes}}
In this supplementary material, we provide additional experiments for both regression and classification. We also provide details on learning the hyperparameters of the RBF kernel, as well as derivations of the proposed acquisition functions. In addition, we give an overview on the changes made to the EP approximation of the GP classifier in order to adapt the model to the setting with weak annotations. 

\section{Additional regression experiments}
We provide additional active learning experiments on the sine curve data set, investigating the effect of cost function, data distribution and of learning the hyperparameters on model performance.

\subsection{Effect of cost function}
\begin{figure}
    \includegraphics[width=\textwidth]{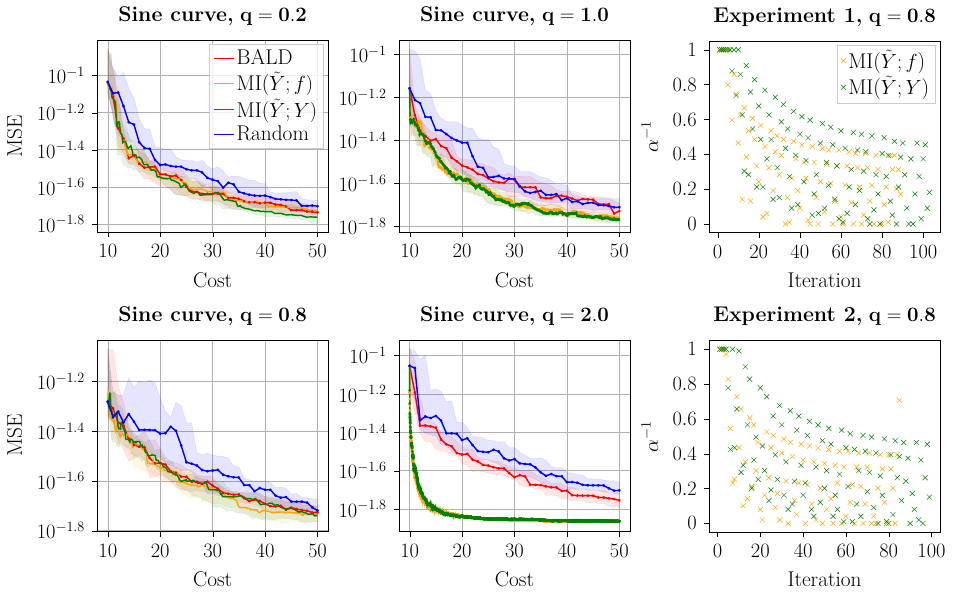}
    \caption{\textit{Left and middle}: Median, first and third quartiles of the test MSE obtained from each set of 15 experiments on the sine curve data set, for different values of the parameter $q$ of the cost function. \textit{Right}: Annotation precisions selected by $\text{MI}(\Noisyy; \model)$ and $\text{MI}(\Noisyy; \Y)$ for $q=0.8$. Shown are the selected precision levels over the course of two experiments (Experiment 1 and Experiment 2, respectively).}
    \label{fig:add_q}
\end{figure}

We investigate the effect of the parameter $q$ of the cost function $C(\alpha)=(1+c\alpha^{-1})^{-q}$, with $c=9.0$, on the behaviour of the active learning algorithm with weak supervision. We run experiments using the sine curve data set with a uniform input distribution and a budget of $B=50$. For calculating the $\text{MI}(\Noisyy;\Y)$ acquisition function, we follow the graphical model in \cref{fig:pgm_noise}(c). Results using $q=0.2, 0.8, 1.0$ and $2.0$ are shown in \cref{fig:add_q}. Most gain in model performance using weak annotations is observed for $q=2.0$, where $\text{MI}(\Noisyy;\model)$ and $\text{MI}(\Noisyy;\model)$ acquires all annotations with the lowest precision. For $q=0.2$, the aforementioned methods instead resort to acquiring a majority of annotations with the highest precision, and give similar model performance to BALD. 

For the cost function with $q = 0.8$, the optimal annotation precision differs over iterations for both $\text{MI}(\Noisyy;\model)$ and $\text{MI}(\Noisyy;\model)$. We show the selected values of $\alpha^{-1}$ over the course of two experiments in the right column of \cref{fig:add_q}. We observe that the algorithms start with selecting annotations of low precision. Thereafter, a pattern can be seen, where the annotation precision is gradually increased. 
We might interpret the general patterns as portraying an initial phase of exploration. Note that the exception seen in experiment 2 (lower right of \cref{fig:add_q}), where $\text{MI}(\Noisyy; \model)$ selects an annotation of low precision just at the end of the experiment, is due to the budget being nearly exhausted.

\subsection{Effect of learning hyperparameters}
\begin{figure}
\centering
      \includegraphics[width=0.8\textwidth]{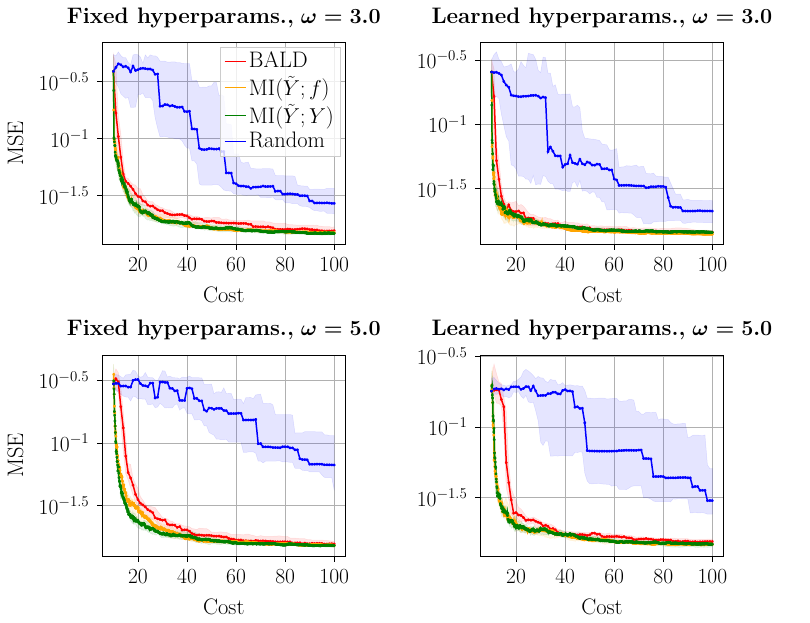}
    \caption{Effect of learning the hyperparameters of the GP kernel in the active learning experiments performed on the sine curve data set. Shown is median, first and third quartiles of the test MSE obtained from each set of 15 experiments.}
    \label{fig:hp_effect}
\end{figure}

We evaluate the effect of learning the hyperparameters of the RBF kernel during the active learning loop, using maximum marginal likelihood estimation. We perform experiments on the sine curve data set using the cost function $C(\alpha)=(1+c\alpha^{-1})^{-q}$, with $c=9.0$ and $q=1.0$, and a budget of $B=100$. Experiments are run with a non-uniform data pool (as described in the main paper) and two sine curve frequencies, $\omega=3.0$ and $\omega=5.0$. Results are shown in \cref{fig:hp_effect}. For the purpose of comparison, we include results obtained when keeping the hyperparameters of the kernels fixed at $a=1.0$ and $l=3.0/\omega$. While learning the hyperparameters does not seem to have a large effect on the conclusions regarding the relative performance of the active learning algorithms, the test MSE is slightly improved overall. 

\subsection{Effect of data distribution}
\begin{figure}
    \includegraphics[width=\textwidth]{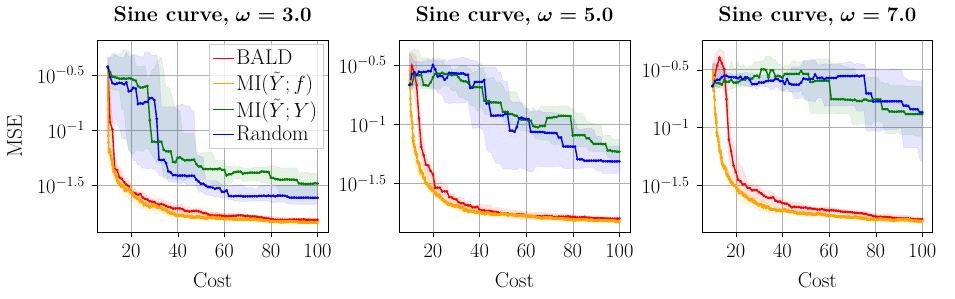}
    \caption{Median, first and third quartiles of the test MSE obtained from each set of 15 experiments on the sine curve data set, for different values of the frequency, $\omega$. It is assumed that $\Noisyy$ is independent of $\model$ and $\X$ given $\Y$.}
    \label{fig:add_omega}
\end{figure}

We provide an additional set of experiments using the sine curve data set, investigating the effect of the sine curve frequency, $\omega$, on the performance of active learning with weak supervision. We assume that $\Noisyy$ is independent of $\model$ and $\X$ given $\Y$ (\cref{fig:pgm_noise}(b)), following the conditional distribution
\begin{align*}
    \Noisyy \mid \Y \sim \gaussian(Y, 0.09/\alpha).
\end{align*}
As in the previous set of experiments, we use the cost function $C(\alpha)=(1+c\alpha^{-1})^{-q}$ with $c= 9.0$ and $q=1.0$ and a budget of $B=100$. We run experiments using the setting of a non-uniform data pool (as described in the main paper) and with frequencies $\omega=3.0, 5.0$ and $7.0$. The lengths scale of the models' RBF kernels are set as $l=3.0/\omega$. Results are shown in \cref{fig:add_omega}. The $\text{MI}(\Noisyy;\model)$ acquisition function improves model performance over BALD in all cases, but particularly for the larger curve frequencies, $\omega=5.0$ and $\omega=7.0$. The $\text{MI}(\Noisyy;Y)$ acquisition function does not work well in this setting, as discussed in the main paper. 

\section{Additional classification experiments}
We investigate the effect of the distribution of the noisy annotation, $\Noisyy$, on the performance of active learning with weak supervision for binary classification. An additional set of experiments are performed using the artificial classification data sets (Version 1, Version 2 and Version 3) but with $\kappa = 0.5$ and $\gamma = 0.5$ in \cref{eq:gpc_prior}, such that $\omega_\alpha \in [0.5, 1.0]$. 

\begin{figure}[b]
     \includegraphics[width=\textwidth]{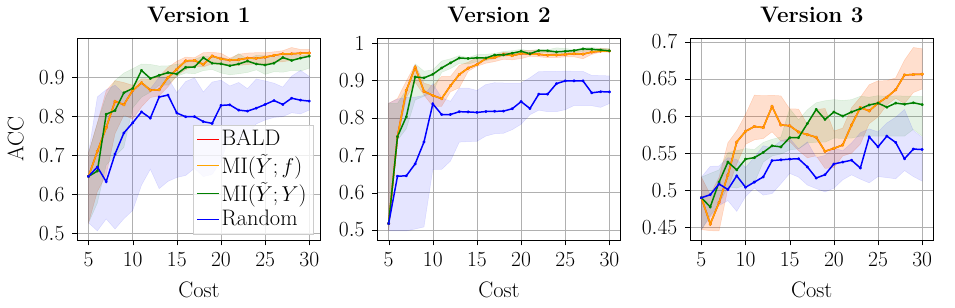}
    \caption{Median, first and third quartiles of the test accuracy obtained from each set of 15 classification experiments. The minimum label flip probability is set to $1 - \kappa = 0.5$.}
    \label{fig:toy_class_exp_2}
\end{figure}

We use the same cost function as in the previous set of classification experiments: $C(\alpha)=0.1+0.9\alpha$. Results are shown in \cref{fig:toy_class_exp_2}. In this setting, $\text{MI}(\Noisyy;\model)$ as well as $\text{MI}(\Noisyy;\Y)$ resort to selecting the highest precision in all cases, rendering the results of the first to be identical to that of BALD. In contrast to the setting with $\kappa=0.8$ and $\gamma = 0.2$, where we observed an advantage with using $\text{MI}(\Noisyy;\model)$ over $\text{MI}(\Noisyy;\Y)$, the performance of $\text{MI}(\Noisyy;\Y)$ is similar to $\text{MI}(\Noisyy;\model)$ in these experiments. 

\section{Details on learning hyperparameters}
We learn the hyperparameters of the RBF kernel of a GP regressor using data $\noisydatatrain{n} = (\xtrain{n}, \noisyytrain{n}, \atrain{n})$, by minimising the marginalised negative log-likelihood
\begin{align*}
    \text{NLL} \propto &\log \big|K\big(\xtrain{n}, \xtrain{n}\big) + \text{diag}\big(\sigma^2(\xtrain{n}) + \gamma\atrain{n}{}^{-1}\big)\big| \\ &+ \noisyytrain{n}{}^\top \left(K\big(\xtrain{n}, \xtrain{n}\big) + \text{diag}\big(\sigma^2(\xtrain{n}) + \gamma\atrain{n}{}^{-1}\big)\right)^{-1} \noisyytrain{n}.
\end{align*}
where $|\cdot|$ denotes the matrix determinant, $K\big(\xtrain{n}, \xtrain{n}\big)$ is a $n\times n$ matrix with elements $K\big(X_i^{(\text{train})}, X_j^{(\text{train})}\big)$ at positions $(i,j)$ and $\text{diag}\big(\sigma^2(\xtrain{n}) + \gamma\atrain{n}{}^{-1}\big)$ is a diagonal matrix of size $n$ with elements $\sigma^2(\X_i^{(\text{train})}) + \gamma / \alpha_i^{(\text{train})}$ at positions $(i, i)$. We use Adam optimisation \citep{Kingma2015} and a learning rate of 0.1. We stop training at epoch $t$ if the current negative log-likelihood, $\text{NLL}_{t}$ has not improved over the previous epoch, according to
\begin{align*}
    \frac{\text{NLL}_{t-1}-\text{NLL}_{t}}{\max(\text{abs}(\text{NLL}_{t-1}, \text{NLL}_{t}, 1))} \leq \epsilon,
\end{align*}
with $\epsilon=5\cdot 10^{-2}$. We also stop training if the current (maximum) gradient is smaller than or equal to $1\cdot 10^{-5}$, or if the maximum of 100 training epochs is reached.

For the UCI data experiments we, in addition to learning the hyperparameters, standardise the input data in each iteration of the active learning loop. 
We additionally train the model to predict the deviation from the target mean, by subtracting the mean of the annotations in the current training set from each acquired annotation.
\section{Acquisition functions for regression}
We provide derivations for the BALD, $\text{MI}(\Noisyy;\model)$ and $\text{MI}(\Noisyy;\Y)$ acquisition functions for regression. Derivations are based on the (differential) entropy of a univariate Gaussian variable, which for a random variable $\Y \sim \mathcal{N}(\mu, \sigma^2)$ is given by (see e.g. \citet{Thomas2006})
\begin{align*}
    \entropy{\Y} & = - \int_{\yspace} \gaussian(y; \mu, \sigma^2) \log \gaussian(y; \mu, \sigma^2)dy = 0.5\big(1 + \log(2\pi) + \log(\sigma^2)\big).
\end{align*}

\subsection{BALD}
The BALD acquisition function is given in \cref{eq:mi_bald}. Using the differential entropy of a Gaussian random variable, and following \cref{eq:regr_noise_model,eq:gp_pred_ws} with $\alpha \rightarrow, \infty$ we have
\begin{align*}
    \entropy{\Y \mid \X, \datatrain{n}}
    = 0.5\big(1 + \log(2\pi) + \log \big(\sigma_{*}^{2} + \sigma^2(X)\big)\big)
\end{align*}
and from \cref{eq:gp_prior}
\begin{align*}
    \expectation_{\model \mid \X, \datatrain{n}}\big[\entropy{\Y \mid \X, \model}\big] = 0.5\big(1 + \log(2\pi) + \log \big(\sigma^2(X)\big)\big).
\end{align*}
Hence,
\begin{align*}
    \text{MI}(\Y;\model \mid \X, \datatrain{n}) = 0.5\big( \log \big(\sigma_{*}^{2} + \sigma^2(X)\big) - \log \big(\sigma^2(X)\big)\big).
\end{align*}

\subsection{MI($\Noisyy;\model$)}
The $\text{MI}(\Noisyy;\model)$ acquisition function is derived similarly to the BALD acquisition function but using a general precision level $\alpha$ in \cref{eq:regr_noise_model,eq:gp_pred_ws}. The numerator in \cref{eq:prop_obj} is given by
\begin{align*}
    \text{MI}(\Y;\model \mid \X, \alpha, \noisydatatrain{n}) = 0.5\big(\log \big(\sigma_{*}^{2} + \sigma^2(X) + \gamma/\alpha \big) - \log \big(\sigma^2(X) + \gamma/\alpha \big)\big).
\end{align*}

\subsection{MI($\Noisyy;\Y$)}
The $\text{MI}(\Noisyy;\Y)$ acquisition function is obtained by replacing $\model$ with $\Y$ in \cref{eq:prop_obj}. The first term of the numerator of this function, is the same as that for the mutual information between $\Noisyy$ and $\model$
\begin{align*}
   \entropy{\Noisyy \mid \X, \alpha, \noisydatatrain{n}} = 0.5\big(1 + \log(2\pi) + \log \big(\sigma_{*}^{2} + \sigma^2(X) + \gamma/\alpha \big)\big).
\end{align*}
The second term of the expression for the mutual information between $\Noisyy$ and $\Y$ will depend on how we model $\Y$ and its relation to $\Noisyy$. We assume that $\Y$ follows the distribution in \cref{eq:regr_noise_model} with $\alpha \rightarrow \infty$. If $\Noisyy$ is independent of $\model$ and $\X$ given $\Y$, as illustrated in \cref{fig:pgm_noise}(b), and
\begin{align*}
    \Noisyy \mid \Y \sim \gaussian \big(\Y, \gamma/\alpha),
\end{align*}
then it follows directly that 
\begin{align*}
    \expectation_{\Y \mid \X, \alpha, \noisydatatrain{n}}\big[\entropy{\Noisyy \mid \X, \alpha, \Y}\big] = 0.5\big(1 + \log(2\pi) + \log \big(\gamma/\alpha \big)\big).
\end{align*}
From this, the conditional mutual information is given by
\begin{align*}
    \text{MI}(\Noisyy;\Y \mid \X, \alpha, \noisydatatrain{n}) = 0.5\big( \log \big(\sigma_{*}^{2} + \sigma^2(X) + \gamma / \alpha \big) - \log \big(\gamma/\alpha \big)\big).
\end{align*}

If $\Noisyy$ is instead independent of $\Y$ given $\model$ and $\X$, as illustrated in \cref{fig:pgm_noise}(c), we can find the second term of the mutual information by first deriving the joint conditional distribution of $\Noisyy$ and $\Y$ from \cref{eq:gp_pred_ws}
\begin{align*}
    \prob(\Noisyy, \Y \mid \X, \alpha) = \gaussian \Big(\begin{bmatrix} \mu_* \\ \mu_* \end{bmatrix}, \begin{bmatrix} \sigma_*^2 + \sigma^2(X) + \gamma / \alpha & \sigma_*^2 \\ \sigma_*^2 & \sigma_*^2 + \sigma^2(X)   \end{bmatrix} \Big).
\end{align*}
From this joint distribution, we find the conditional distribution of $\Noisyy$ as
\begin{align*}
    \prob(\Noisyy \mid \X, \alpha, \Y) = \gaussian \bigg(\mu_* + \frac{\sigma_*^2(Y - \mu_*)}{\sigma_*^2 + \sigma^2(X)}, \quad \sigma_*^2 + \sigma^2(X) + \gamma / \alpha - \frac{\sigma_*^4}{\sigma_*^2 +\sigma^2(X)}\bigg).
\end{align*}
Hence,
\begin{align*}
    &\expectation_{\Y \mid \X, \noisydatatrain{n}}\big[\entropy{\Noisyy \mid \X, \alpha, \Y}\big] = \\ & \qquad \qquad 0.5 \bigg(1 + \log(2\pi)  + \log \Big(\sigma_*^2 + \sigma^2(X) + \gamma / \alpha - \frac{\sigma_*^4}{\sigma_*^2 + \sigma^2(X)} \Big)\bigg).
\end{align*}
The numerator of \cref{eq:prop_obj}, replacing $\model$ with $\Y$, is then given by
\begin{align*}
     &\text{MI}(\Noisyy;\Y \mid \X, \alpha, \noisydatatrain{n}) = \\ & \qquad 0.5\bigg(\log \big(\sigma_{*}^{2} + \sigma^2(X) + \gamma / \alpha \big) - \log \Big(\sigma_*^2 + \sigma^2(X) + \gamma / \alpha - \frac{\sigma_*^4}{\sigma_*^2 + \sigma^2(X)} \Big) \bigg).
\end{align*}

\section{Acquisition functions for classification}
What follows are derivations of the proposed $\text{MI}(\Noisyy;\model)$ acquisition function as well as the $\text{MI}(\Noisyy;\Y)$ and BALD acquisition functions for classification. In binary classification, the aforementioned acquisition functions are based on the mutual information between two Bernoulli random variables. We use Shannon entropy (following \citet{Houlsby2011}), which, for a Bernoulli random variable, is given by
\begin{align*}
    h(p) = -p\logt(p) - (1-p)\logt(1-p).
\end{align*}
%
We assume that the posterior over $\model$ is approximated with a Gaussian distribution $\mathcal{N}(\mu_*, \sigma_*^2)$.

\subsection{BALD}

Following \cite{Houlsby2011}, using a Gaussian approximation to the posterior of $\model$, the first term in the expression for mutual information given in \cref{eq:mi_bald} is
\begin{align*}
    \entropy{\Noisyy \mid X, \datatrain{n}} \approx h\Bigg( \Phi \Big( \frac{\mu_*}{\sqrt{\sigma_*^2 + 1}} \Big) \Bigg)
\end{align*}
The following Taylor expansion of order three 
around $\model = 0$ is used to approximate the second term of the mutual information
\begin{align*}
    \log (h(\Phi(f))) \approx -\frac{\model^2}{\pi\log(2)},
\end{align*}
according to \cite{Houlsby2011}.
Using this approximation, the following expression is obtained
\begin{align*}
    \expectation_{\model \mid \X, \datatrain{n}} [\entropy{Y \mid X, \model}] \approx \frac{C}{\sqrt{\sigma_*^2 + C^2}} \exp{\left(-\frac{\mu_*^2}{2(\sigma_*^2 + C^2)}\right)}, 
\end{align*}
where $C = \sqrt{\pi \log(2)/2}$.

The final acquisition function is 
\begin{align*}
    \text{MI}(\Y;\model \mid \X, \datatrain{n}) \approx h\Bigg(\Phi\Big(\frac{\mu_*}{\sqrt{\sigma_*^2 + 1}}\Big) \Bigg) -
\frac{C}{\sqrt{\sigma_*^2 + C^2}} \exp\Big(-\frac{\mu_*^2}{2(\sigma_*^2 + C^2)}\Big),
\end{align*}
with $C = \sqrt{\pi \log(2)/2}$.

\subsection{MI($\Noisyy;\model$)}
To derive the $\text{MI}(\Noisyy;\model)$ acquisition function for the binary GP classifier, first note that the noisy target variable $\Noisyy$, with precision $\alpha$, follows a Bernoulli distribution conditioned on $\model(X)$
\begin{align*}
    \Noisyy \mid \model, \X, \alpha \sim \text{Bernoulli}(\omega_\alpha \Phi(f(X)) + (1-\omega_\alpha)(1 - \Phi(f(X)))),
\end{align*}
where $\omega_\alpha = \kappa + \gamma \alpha$.
Approximating the posterior over $\model$ with a Gaussian distribution $\mathcal{N}(\mu_*, \sigma_*^2)$, the first term in the expression for the mutual information is given by
\begin{align*}
    \entropy{\Noisyy \mid \X, \alpha, \noisydatatrain{n}} \approx h\Big(\int(\omega_\alpha \Phi(f) + (1-\omega_\alpha)(1 - \Phi(f))) \gaussian \big(f; \mu_*, \sigma_*^2 \big)df\Big),
\end{align*}
where the approximation follows from the Gaussian approximation to the posterior over $\model$. The integral can be solved directly 
\begin{align*}
    &\int(\omega_\alpha \Phi(f) + (1-\omega_\alpha)(1 - \Phi(f))) \gaussian \big(f; \mu_*, \sigma_*^2 \big)df = \\ & \qquad (2\omega_\alpha - 1)\int \Phi(f) \gaussian \big(f; \mu_*, \sigma_*^2 \big)df + 1 - \omega_\alpha = (2\omega_\alpha - 1)\frac{\mu_*}{\sqrt{\sigma_*^2 + 1}} + 1 - \omega_\alpha. 
\end{align*}

For the second term of the mutual information, we have
\begin{align*}
    \expectation_{\model \mid X, \alpha, \noisydatatrain{n}} [\entropy{\Noisyy \mid X, \alpha \model}] = \int h((2\omega_\alpha-1)\Phi(\model)+1-\omega_\alpha) \gaussian (\model \mid \mu_*, \sigma_*^2)d\model.
\end{align*}
Similar to deriving the BALD acquisition function, we need to apply some approximation in order to solve this integral. Following \citep{Houlsby2011} we approximate the integral using a Taylor expansion, but this time of the function
\begin{align*}
    g(x) = \log \big( h((2\omega_\alpha - 1)\Phi(x) + 1 - \omega_\alpha) - h(\omega_\alpha) \big)
\end{align*}
around $x=0$.
We will do a Taylor expansion of order three 
\begin{align*}
    g(x) = g(0) + g'(0)x + \frac{g''(0)}{2}x^2 + \frac{g^{(3)}(0)}{6}x^3 + \mathcal{O}(x^4). 
\end{align*}
The first term is given by
\begin{align*}
    g(0) = \log \big( h((2\omega_\alpha - 1)\cdot \Phi(0) + 1 - \omega_\alpha - h(\omega_\alpha)) \big)  = \log(1 - h(\omega_\alpha)),
\end{align*}
and the first derivative follows as
\begin{align*}
    &g'(x) = \\ & \quad \frac{(2\omega_\alpha - 1)\Phi'(x) \big(\logt((1-2\omega_\alpha)\Phi(x) + \omega_\alpha) - \logt((2\omega_\alpha - 1)\Phi(x) + 1 - \omega_\alpha)\big)}{g(x)}, \\
    &g'(0) = 0.
\end{align*}
For the second derivative, we have
\begin{align*}
    &g''(x) = \\ & \quad -g'(x)^2 + \frac{1}{\log(2) g(x)} \bigg(- \frac{(2\omega_\alpha-1)^2\Phi'(x)^2}{\big((1 - 2\omega_\alpha)\Phi(x) + \omega_\alpha \big)\big((2\omega_\alpha - 1)\Phi(x) + 1 - \omega_\alpha\big)} \\ & \quad + \Phi''(x) (2\omega_\alpha-1)(\log((1-2\omega_\alpha)\Phi(x) + \omega_\alpha) - \log((2\omega_\alpha - 1)\Phi(x) + 1 - \omega_\alpha) \bigg), \\
    & g''(0)  = \frac{-2(2\omega_\alpha - 1)^2}{\pi \log(2) (1 - h(\omega_\alpha))}.
\end{align*}
Finally,
\begin{align*}
    g(x) = \log\big(1 - h(\omega_\alpha)\big) - \frac{(2\omega_\alpha - 1)^2}{\pi \log(2) (1 - h(\omega_\alpha))}x^2 + \mathcal{O}(x^4).
\end{align*}
The approximation is up to $\mathcal{O}(x^4)$, as the function is even and, hence, we will find that the $x^3$ term in the expansion will be 0. 
Following the Taylor expansion, we can use the following approximation to the conditional entropy of $\Noisyy$ 
\begin{align*}
    h((2\omega_\alpha - 1)\Phi(\model) + 1 - \omega_\alpha) \approx (1-h(\omega_\alpha))\exp\bigg(\frac{-(2\omega_\alpha - 1)^2}{\pi \log(2) (1 - h(\omega_\alpha))} x^2 \bigg) + h(\omega_\alpha).
\end{align*}
This approximation, together with the Gaussian approximation to the posterior, gives
\begin{align*}
     &\expectation_{\model \mid \X, \alpha, \noisydatatrain{n}} \big[\entropy{\Noisyy \mid \X, \alpha, \model}\big] \approx   \int h((2\omega_\alpha-1)\Phi(\model)+1-\omega_\alpha) \gaussian(\model \mid \mu_*, \sigma_*^2)d\model \\& \qquad \qquad \approx (1-h(\omega_\alpha))\int \exp\bigg(\frac{-2(2\omega_\alpha - 1)^2}{\pi \log(2) (1 - h(\omega_\alpha))}\bigg) \gaussian(\model \mid \mu_*, \sigma_*^2) d\model + h(\omega_\alpha) \\& \qquad \qquad = \frac{(1-h(\omega_\alpha))}{\sqrt{1 + 2C\sigma_*^2}}\exp\bigg(-\frac{C \mu_*^2}{1 + 2C\sigma_*^2} \bigg) + h(\omega_\alpha), 
\end{align*}
where $C = (2\omega_\alpha - 1)^2(\pi \log(2) (1 - h(\omega_\alpha)))^{-1}$.

Using the aforementioned approximation, the numerator of the $\text{MI}(\Noisyy;\model)$ acquisition function in \cref{eq:prop_obj} is given by
\begin{align*}
     &\text{MI}(\Noisyy;\model \mid \X, \alpha, \noisydatatrain{n}) \\ & \qquad \qquad \approx h\Bigg( \Phi \Big( \frac{\mu_*}{\sqrt{\sigma_*^2 + 1}} \Big) \Bigg) - \frac{(1-h(\omega_\alpha))}{\sqrt{1 + 2C\sigma_*^2}}\exp\bigg(-\frac{C \mu_*^2}{1 + 2C\sigma_*^2} \bigg) + h(\omega_\alpha),
\end{align*}
with $C = (2\omega_\alpha - 1)^2(\pi \log(2) (1 - h(\omega_\alpha)))^{-1}$.

\subsection{MI($\Noisyy; \Y$)}
The first term in the conditional mutual information between $\Noisyy$ and $\Y$ is the same as for the MI($\Noisyy; \model$) acquisition function. Hence, we only need to derive the second term which can be simplified according to 
\begin{align*}
    &\expectation_{\Y \mid \X, \alpha, \noisydatatrain{n}}\big[\entropy{\Noisyy \mid \X, \omega_\alpha, \Y}\big] = \\ & \qquad \qquad  \Phi\Big(\frac{\mu_*}{\sqrt{1 + \sigma_*^2}}\Big) h(\omega_\alpha) + (1-\Phi\Big(\frac{\mu_*}{\sqrt{1 + \sigma_*^2}}\Big)) h(1 - \omega_\alpha) = h(\omega_\alpha),
\end{align*}
where the last equality follows as $h(\omega_\alpha) = h(1 - \omega_\alpha)$.

The numerator of the acquisition function, exchanging $\model$ with $\Y$ in \cref{eq:prop_obj}, is given by
\begin{align*}
     \text{MI}(\Noisyy;\Y \mid \X, \alpha, \noisydatatrain{n}) \approx h\Bigg( \Phi \Big( \frac{\mu_*}{\sqrt{\sigma_*^2 + 1}} \Big) \Bigg) - h(\omega_\alpha).
\end{align*}
The expression is approximate following the Gaussian approximation to the posterior.

\section{EP for GP classifier with weak supervision}

In the classification experiments, the EP algorithm is used to fit the approximate Gaussian posterior to the training data. \citet{Rasmussen2006} divide each iteration of the algorithm into four steps, where, for each observation, the following steps are performed: 

\begin{enumerate}
    \item Derive a marginal cavity distribution from the approximate posterior, leaving out the approximate likelihood of observation $i$.
    \item Obtain the non-Gaussian marginal by combining the cavity distribution with the exact likelihood.
    \item Find a Gaussian approximation to the marginal.
    \item Update the approximate likelihood of observation $i$ to fit the Gaussian approximation from the previous step.
\end{enumerate}
In what follows, we will derive the equations relevant for updating the posterior marginal moments (step 3) for the GP classifier with weak supervision. The other steps remain the same as for the standard GP classifier, and we therefore refer to \citep{Rasmussen2006} for details.

\subsection{Marginal moments with weak supervision}

We derive the moments of the non-Gaussian marginal posterior of $\model_i$ obtained in step 2 of the EP algorithm, which in turn are used to update the approximate posterior in step 4. First, note that the probability distribution of $\Noisyy$ given $\model$ can be derived by marginalising over $\Y$ to get
\begin{align*}
    \prob(\Noisyy=\noisyy_i \mid \model=\model_i) = (2\omega_\alpha-1)\Phi(\noisyy_i\model_i) + 1 - \omega_\alpha.
\end{align*}
Let $\mucavity$ and $\sigmacavity^2$ denote the mean and variance, respectively, of the cavity distribution obtained in step 1 of the EP algorithm. Then, the marginal of $\model_i$ has the form
\begin{align*}
    q(f_i) = \Tilde{Z}^{-1} \big((2\omega_\alpha - 1)\Phi (f_i\noisyy_i) + (1-\omega_\alpha)\big)\gaussian(f_i; \mucavity, \sigmacavity^2)
\end{align*}
with
\begin{align*}
   &\tilde{Z} = (2\omega_\alpha - 1)Z + 1-\omega_\alpha, \\
   & Z = \fullint \Phi (f_i\noisyy_i) \gaussian(f_i;\mucavity, \sigmacavity^2)df_i, \\ 
    &\Phi(x) = \int_{-\infty}^x \gaussian(v; 0, 1)dv.
\end{align*}
We will start with deriving the normalising constant, $\Tilde{Z}$. First, to obtain an expression for $Z$, let $z = v - f_i + \mucavity$ and $w = f_i -\mu$. Assume $\noisyy_i=1$, then
\begin{align*}
    &Z_{\noisyy_i=1} = \frac{1}{2\pi\sigmacavity} \fullint \int_{-\infty}^{f_i} \exp\Big(-\frac{v^2}{2} -\frac{(f_i - \mucavity)^2}{2\sigmacavity^2} \Big) dvdf_i \\ & \qquad = 
    \frac{1}{2\pi\sigmacavity} \int_{\infty}^{\mucavity} \fullint  \exp\Big(-\frac{(z+w)^2}{2} -\frac{w^2}{2\sigmacavity^2} \Big)dwdz \\ & \qquad = \frac{1}{2\pi\sigmacavity}\int_{\infty}^{\mucavity} \fullint \gaussian \Bigg(\begin{bmatrix} w \\ z\end{bmatrix}; \begin{bmatrix} 0 \\ 0\end{bmatrix}, \begin{bmatrix} \sigmacavity^2 & -\sigmacavity^2  \\ -\sigmacavity^2 & 1 + \sigmacavity^2 \end{bmatrix} \Bigg)dwdz 
\end{align*}
The inner integral is a marginalisation over $w$
\begin{align*}
     Z_{\noisyy_i=1} &= \frac{1}{\sqrt{2\pi(1+\sigmacavity^2)}}\int_{\infty}^{\mucavity}  \exp\Bigg(-\frac{z^2}{2(1 + \sigmacavity^2)} \Bigg)dz  = \Phi\Bigg(\frac{\mucavity}{\sqrt{1 + \sigmacavity^2}}\Bigg).
\end{align*}
For $\noisyy_i=-1$, use $\Phi(-x) = 1 - \Phi(x)$ to get
\begin{align*}
    Z_{\noisyy_i=-1} = 1 - \Phi\Bigg(\frac{\mucavity}{\sqrt{1 + \sigmacavity^2}}\Bigg) = \Phi\Bigg(-\frac{\mucavity}{\sqrt{1 + \sigmacavity^2}}\Bigg).
\end{align*}
Taken together, we see that
\begin{align*}
    Z = \Phi\Bigg(\frac{\noisyy_i\mucavity}{\sqrt{1 + \sigmacavity^2}}\Bigg).
\end{align*}
We obtain the normalisation constant, $\tilde{Z}$, as
\begin{align*}
     \Tilde{Z} &= \fullint \big((2\omega_\alpha - 1)\Phi (f_i\noisyy_i) + (1-\omega_\alpha)\big) \gaussian(f_i;\mucavity, \sigmacavity^2)df_i \\ &=
     (2\omega_\alpha -1 ) Z + (1-\omega_\alpha)\fullint \gaussian(f_i;\mucavity, \sigmacavity^2)df_i \\ &= (2\omega_\alpha -1 ) Z + (1-\omega_\alpha)
\end{align*}

Next, to compute the moments of $q(f_i)$, take the derivative of $\Tilde{Z}$ with respect to $\mucavity$
\begin{align*}
    \frac{\partial \Tilde{Z}}{\partial \mucavity} &= \fullint \frac{f_i - \mucavity}{\sigmacavity^2}\big((2\omega_\alpha - 1)\Phi (f_i\noisyy_i) + (1-\omega_\alpha)\big)\gaussian(f_i; \mucavity, \sigmacavity^2)df_i \\ &= \fullint \frac{f_i - \mucavity}{\sigmacavity^2}\Tilde{Z}q(f_i)df_i \\ &= \frac{\partial}{\partial \mucavity} (2\omega_\alpha - 1)\Phi(z) + 1-\omega_\alpha,
\end{align*}
with
\begin{align*}
    &\frac{\partial}{\partial \mucavity} \big((2\omega_\alpha - 1)\Phi(z) + 1-\omega_\alpha\big) = (2\omega_\alpha - 1) \gaussian(z; 0, 1) \frac{\partial z}{\partial\mucavity} \\ & \qquad \qquad =  (2\omega_\alpha - 1) \gaussian(z; 0, 1) \frac{\noisyy_i}{\sqrt{1+\sigmacavity^2}}.
\end{align*}
Hence, we have
\begin{align*}
    \frac{\Tilde{Z}}{\sigmacavity^2}\int f_i q(f_i)df_i - \frac{\mucavity}{\sigmacavity^2} \Tilde{Z} = (2\omega_\alpha - 1) \gaussian(z; 0, 1) \frac{\noisyy_i}{\sqrt{1+\sigmacavity^2}}
\end{align*}
and therefore
\begin{align*}
    \hat{\mu}_i = \int f_i q(f_i)df_i = \mucavity - \sigmacavity^2(2\omega_\alpha - 1) \gaussian(z; 0, 1) \frac{\noisyy_i}{\Tilde{Z}\sqrt{1+\sigmacavity^2}}.
\end{align*}
For the second moment, take the second derivative of $\Tilde{Z}$  with respect to $\mucavity$
\begin{align*}
    &\frac{\partial^2 \Tilde{Z}}{\partial \mucavity^2} = \fullint \Big(\frac{f_i^2}{\sigmacavity^4} - \frac{2\mucavity f_i}{\sigmacavity^2} + \frac{\mucavity^2}{\sigmacavity^2} - \frac{1}{\sigmacavity^2}\Big)\Tilde{Z}q(f_i)df_i \\ & \qquad= \frac{\partial^2}{\partial \mucavity^2} (2\omega_\alpha - 1)\Phi(z) + 1-\omega_\alpha,
\end{align*}
with
\begin{align*}
    &\frac{\partial^2}{\partial \mucavity^2} (2\omega_\alpha - 1)\Phi(z) + 1-\omega_\alpha = (2\omega_\alpha - 1)\frac{\noisyy_i}{\sqrt{1+\sigmacavity}} \frac{\partial}{\partial \mucavity}\gaussian(z;0,1) \\ & \qquad \qquad = - z\gaussian(z;0,1) \frac{(2\omega_\alpha - 1)}{1 + \sigmacavity^2}.
\end{align*}
Following this,
\begin{align*}
    &\frac{\Tilde{Z}}{\sigmacavity^4}\fullint f_i^2 q(f_i)df_i - \frac{2\Tilde{Z}\mucavity}{\sigmacavity^4}\fullint f_i q(f_i)df_i + \Tilde{Z}\Big(\frac{\mucavity^2}{\sigmacavity^4} - \frac{1}{\sigmacavity^2}\Big) \\ & \qquad \qquad =  - z\gaussian(z;0,1) \frac{(2\omega_\alpha - 1)}{1 + \sigmacavity^2}
\end{align*}
and
\begin{align*}
    \expectation_{f_i}[f_i^2] = \fullint f_i^2 q(f_i)df_i = 2\mucavity\expectation_{f_i}[f_i] - \mucavity^2 + \sigmacavity^2 - z\gaussian(z; 0, 1)\frac{\sigmacavity^4(2\omega_\alpha - 1)}{\Tilde{Z}(1 + \sigmacavity^2)}.
\end{align*}
We can derive the variance of the Gaussian approximation by
\begin{align*}
    &\hat{\sigma}_i^2 = \expectation_{f_i}[f_i^2] - \hat{\mu}_i^2 =
    \sigmacavity^2 - \gaussian(z;0,1)\frac{\sigmacavity^4 (2\omega_\alpha - 1)}{\Tilde{Z}(1+\sigmacavity^2)}\Big(z + \gaussian(z;0,1)\frac{2\omega_\alpha - 1}{\Tilde{Z}}\Big).
\end{align*}

To summarise, the mean and variance of the Gaussian approximation to the marginal are
\begin{align*}
    \hat{\mu}_i = \int f_i q(f_i)df_i = \mucavity - \sigmacavity^2(2\omega_\alpha - 1) \gaussian(z; 0, 1) \frac{\noisyy_i}{\Tilde{Z}\sqrt{1+\sigmacavity^2}}
\end{align*}
and
\begin{align*}
    \hat{\sigma}_i^2 = \sigmacavity^2 - \gaussian(z;0,1)\frac{\sigmacavity^4 (2\omega_\alpha - 1)}{\Tilde{Z}(1+\sigmacavity^2)}\Big(z + \gaussian(z;0,1)\frac{2\omega_\alpha - 1}{\Tilde{Z}}\Big).
\end{align*}

\end{document}